\documentclass[10pt,twocolumn,letterpaper]{article}

\usepackage{iccv}
\usepackage{times}
\usepackage{epsfig}
\usepackage{graphicx}
\usepackage{amsmath}
\usepackage{amssymb}
\usepackage{booktabs}
\usepackage{array}
\usepackage{multirow}
\usepackage{color}
\usepackage{colortbl}
\usepackage{framed}
\usepackage{bm}
\usepackage{bbm}
\usepackage{xspace}
\usepackage{enumitem}
\usepackage{cite}
\usepackage{xparse}
\usepackage{xcolor}
\usepackage{algorithm}
\usepackage{lipsum}
\usepackage{listings}
\usepackage{url}
\usepackage[accsupp]{axessibility}

\usepackage[pagebackref=true,breaklinks=true,letterpaper=true,colorlinks,bookmarks=false]{hyperref}

\definecolor{Gray}{gray}{0.9}
\definecolor{LightCyan}{rgb}{0.88,0.95,1}
\definecolor{blond}{rgb}{0.98, 0.94, 0.75}

\def \ie {\emph{i.e.}}
\def \eg {\emph{e.g.}}
\def \etal {\emph{et al.}}
\def \softmax {\text{softmax}}
\def \tr {^\intercal}

\newcommand{\tit}[1]{\smallbreak\noindent\textbf{#1.}}
\newcommand{\tinytit}[1]{\noindent\textbf{#1.}}
\newcommand*{\qed}{\null\nobreak\hfill\ensuremath{\square}}

\newcommand{\ours}{PMA-Net\xspace}

\iccvfinalcopy 

\def\iccvPaperID{****} 

\ificcvfinal\fi

\begin{document}

\title{With a Little Help from your own Past: \\Prototypical Memory Networks for Image Captioning}

\author{Manuele Barraco$^1$ \quad Sara Sarto$^1$ \quad Marcella Cornia$^1$ \quad Lorenzo Baraldi$^1$ \quad Rita  Cucchiara$^{1,2}$  \\
$^1$University of Modena and Reggio Emilia, Modena, Italy \quad $^2$IIT-CNR, Pisa, Italy\\
{\tt\small \{name.surname\}@unimore.it}
}

\maketitle
\ificcvfinal\thispagestyle{empty}\fi

\begin{abstract}
Image captioning, like many tasks involving vision and language, currently relies on Transformer-based architectures for extracting the semantics in an image and translating it into linguistically coherent descriptions. Although successful, the attention operator only considers a weighted summation of projections of the current input sample, therefore ignoring the relevant semantic information which can come from the joint observation of other samples. In this paper, we devise a network which can perform attention over activations obtained while processing other training samples, through a \textit{prototypical memory} model. Our memory models the distribution of past keys and values through the definition of prototype vectors which are both discriminative and compact. Experimentally, we assess the performance of the proposed model on the COCO dataset, in comparison with carefully designed baselines and state-of-the-art approaches, and by investigating the role of each of the proposed components. We demonstrate that our proposal can increase the performance of an encoder-decoder Transformer by 3.7 CIDEr points both when training in cross-entropy only and when fine-tuning with self-critical sequence training. Source code and trained models are available at: {\small\url{https://github.com/aimagelab/PMA-Net}}.
\end{abstract}

\section{Introduction}
\label{sec:intro}
Connecting vision and natural language via descriptive expressions is a fundamental human capability and its replication represents a crucial step towards machine intelligence, with applications that range from better human-machine interfaces~\cite{li2021robotic} to accessibility~\cite{gurari2020captioning}. The task of image captioning~\cite{vinyals2015show,xu2015show,karpathy2015deep}, which defines such capability, requires an algorithm to describe a visual input with a natural language sentence. As such, it features unique challenges that span from a grounded and detailed understanding of the visual input~\cite{sennrich2015neural,cornia2019show}, to the selection of visual objects and semantics that are worth mentioning~\cite{li2022comprehending} and their translation into a fluent and coherent sentence.

Image captioning architectures comprise an image encoding part and a language generation approach~\cite{stefanini2021show} and focus on developing appropriate connections between the visual and textual modality. Examples of such innovations include the usage of attentive-like structures~\cite{anderson2018bottom,pan2020x}, the incorporation of attributes~\cite{li2020oscar,you2016image}, objects~\cite{anderson2018bottom,yao2017incorporating} or scene graphs~\cite{yang2019auto,yao2018exploring}. Regardless of the specific approach used to connect the two modalities, though, almost all of the works developed in the last years share the usage of the Transformer architecture~\cite{vaswani2017attention}. Such architecture, indeed, is a natural choice for the task, as it can connect two modalities thanks to its encoder-decoder design and the cross-attention operator, and provides unprecedented performance in sequence and set modeling and generation~\cite{devlin2018bert,radford2019language,cornia2020meshed}.

\begin{figure}[t]
\begin{center}
\includegraphics[width=\linewidth]{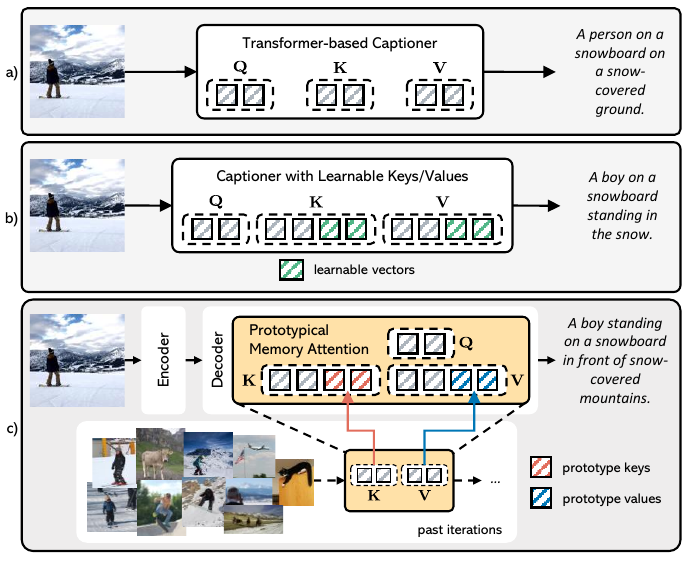}
\end{center}
\vspace{-.2cm}
\caption{Comparison between (a) a standard Transformer-based captioner; (b) a captioner with learnable memory vectors~\cite{cornia2020meshed} and (c) our prototypical memory network.}
\label{fig:firstpage}
\vspace{-.4cm}
\end{figure}

One of the key properties of attention layers is that the output is computed as a weighted combination of linear projections of the inputs. While this provides an ideal context for both visual understanding and sequence generation, it also leaves the door open for injecting relevant information which can not be directly inferred from the current input sample. An interesting attempt in this direction has been made by the Meshed-Memory architecture~\cite{cornia2020meshed}, which proposed to insert additional learnable key/value vectors in the visual encoder with the objective of integrating a-priori knowledge. While successful, learnable vectors can just store information that is useful for the entire training set and do not really act as a ``memory'' of past training items. Instead, having access to past training samples at generation time can be a powerful source of information that can ultimately increase the description quality. For instance, given an input image representing a boy snowboarding in a mountain landscape, a model which has access to other training items might retrieve similar images containing boy, snowboard, and mountains even in a different context, and employ this knowledge in a compositional manner to aid the generation of a correct and fluent sentence (Fig.~\ref{fig:firstpage}).

Following this insight, we devise a \textit{prototypical memory} network, which can recall and exploit past activations generated during training. Our memory is built upon network activations obtained during recent training iterations so that the network has access to a vast set of activations produced while processing other samples from the dataset. In this sense, the memory represents \textit{past knowledge processed by the network itself}.
From the point of view of the architectural design, our memory is fully integrated into attention layers through the addition of keys and values which represents activations from the memory.
To the best of our knowledge, this is the first attempt of integrating a memory of past training items into an image captioning network.

The key element of our proposal is the computation of ``prototypes'' from a bank of past activations, which are obtained by modeling the manifold of past activations and clustering its content in a memory with a given fixed size. This is done both at key and value level, by exploiting the mapping between corresponding keys and values. We further justify our prototype generation approach by investigating the resulting attention distribution from a theoretical point of view.
Experimentally, we assess the performances of the proposed design on the COCO dataset for image captioning, in comparison with state-of-the-art approaches and carefully-designed ablations to study the role of each component of the proposal. Further, we conduct experiments on the nocaps dataset~\cite{agrawal2019nocaps} for novel object captioning and on the robust COCO split~\cite{lu2018neural} for object hallucination.

We believe that the proposed approach can shed light on the effectiveness of employing the space of training items as an additional input to the network, which is currently under-explored and could be in principle applied outside of image captioning. To sum up, the main contribution of this work is the proposal of a prototype memory network for image captioning, which elegantly integrates past activations as an additional input of attention layers. Extensive experiments on COCO, nocaps, and the robust COCO split demonstrate the effectiveness of the proposed approach.

\section{Related Work}
\label{sec:related}
\tinytit{Image Captioning}
Early captioning approaches were constructed by filling pre-defined templates after detecting relevant objects inside of the image~\cite{socher2010connecting,yao2010i2t}. The advent of deep learning then made the use of RNNs and LSTMs a popular choice for the task, and in analogy to the sequence modeling used in machine translation~\cite{sutskever2014sequence}, the basic RNN-based encoder-decoder scheme was employed in conjunction with CNNs for encoding the visual content~\cite{vinyals2015show,rennie2017self,landi2021working}. This approach was later augmented with the addition of attention~\cite{xu2015show,lu2017knowing,you2016image}. Nowadays, attentive and Transformer-based architectures~\cite{vaswani2017attention,dosovitskiy2021image,touvron2021training,radford2021learning} are often employed both in the visual encoding stage, usually applied to refine features from a CNN~\cite{zhang2021rstnet} or ViT~\cite{dosovitskiy2021image}, and as language models~\cite{herdade2019image,cornia2020meshed}. Other solutions~\cite{luo2021dual,nguyen2022grit}, instead, exploit self-attention to effectively combine visual features coming from multiple backbones (\ie~typically a CNN and an object detector), in some cases also finetuning the visual backbones to improve final performances~\cite{nguyen2022grit}.

The introduction of Transformer-based models in image captioning has also brought to the development of effective variants of the self-attention operator~\cite{huang2019attention,herdade2019image,pan2020x,cornia2020meshed,guo2020normalized,liu2020prophet} and to that of vision-and-language early-fusion architectures~\cite{li2020oscar,hu2021scaling,zhou2020unified} based on BERT-like models~\cite{devlin2018bert}. Recently, a common strategy is that of employing visual features extracted from large-scale cross-modal architectures~\cite{shen2022much,mokady2021clipcap,cornia2022universal,barraco2022camel} like CLIP~\cite{radford2021learning}. As done in~\cite{li2022comprehending} and other contemporary works~\cite{sarto2022retrieval,cornia2022universal,ramos2023smallcap}, these multimodal architectures can also enable the enrichment of predicted textual sentences, by means of retrieval components that can be added to the captioning model.

\tit{Memory-Augmented Transformers}
External memories have been used in different ways in Transformer-based architectures, mainly in NLP. Khandelwal~\etal~\cite{khandelwal2019generalization} constructed a memory for language generation as a large table of (key, token) pairs, while Sukhbaatar~\etal~\cite{sukhbaatar2019augmenting} replace feed-forward layers with differentiable memory slots. Recently, Wu~\etal~propose a Memorizing Transformer~\cite{wu2022memorizing} architecture in which they retrieve activations produced over long documents. In vision-and-language, learnable external memories have been successfully employed for image captioning~\cite{cornia2020meshed}, visual relationship recognition~\cite{chen2022reltransformer}, and story generation~\cite{xue2022mmt}.

\begin{figure*}[t]
\begin{center}
\includegraphics[width=0.98\linewidth]{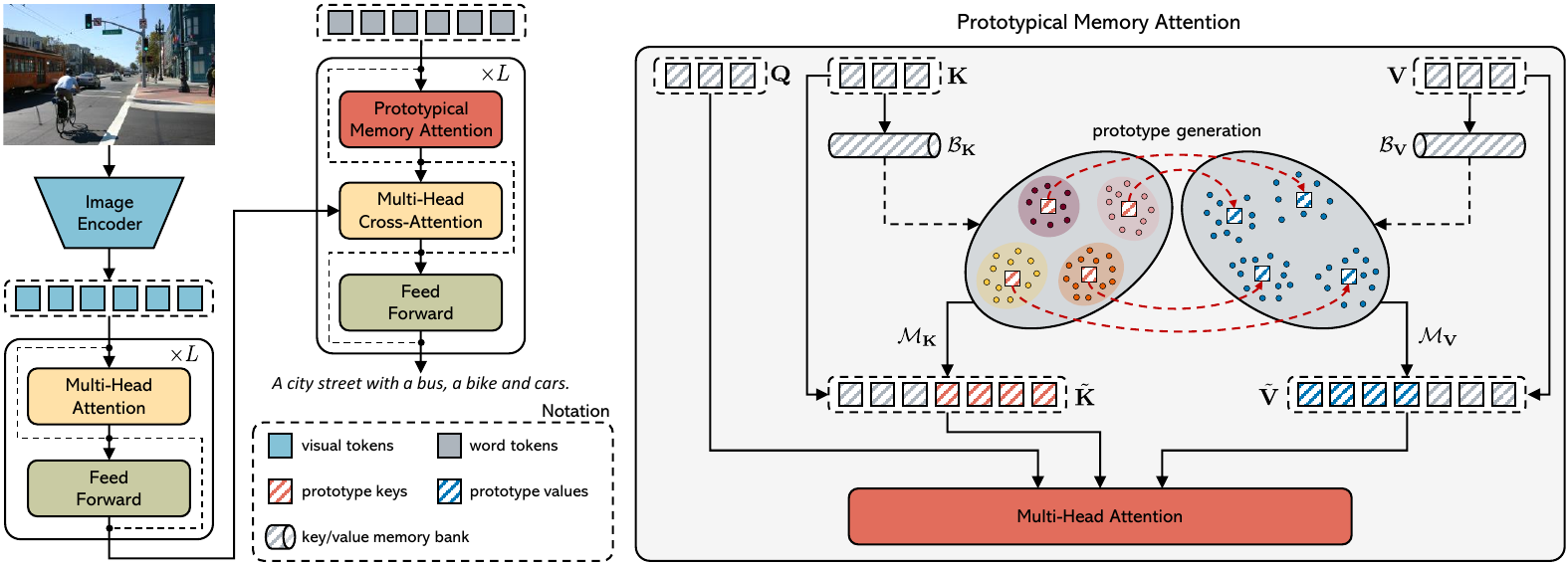}
\end{center}
\vspace{-.2cm}
\caption{Overview of our approach with Prototypical Memory Attention.}
\label{fig:model}
\vspace{-.4cm}
\end{figure*}

\tit{Summary}
Our proposal is different from all the aforementioned approaches, as it employs network activations coming from other training samples as a source of additional knowledge. This research direction might also be seen as an alternative to retrieving information from external knowledge bases, as it represents additional knowledge through network activations rather than with raw data, and further devises a compression step so that the additional information can be stored directly into the network and avoid the usage of external databases.

\section{Proposed Method}
\label{sec:method}
\subsection{Preliminaries: Memory-augmented Attention}
Attention layers operate on triplets of queries, keys and values $(\mathbf{Q}, \mathbf{K}, \mathbf{V})$, which are obtained by linearly projecting items from the same input sequence (self-attention) or from a pair of different input sequences (cross-attention). We are interested in breaking the constraint of operating exclusively on input-dependent data and letting the attention operator consider quantities which are not derived from the current input~\cite{cornia2020meshed,wu2022memorizing}. In memory-augmented attention~\cite{cornia2020meshed}, this is achieved by extending the set of keys and values to include additional memory vectors. As a result, the attention operation can employ both input-dependent and memory-specific keys and values, as follows:
\begin{equation}
    \label{eq:memory_def}
    \begin{gathered}
        \tilde{\mathbf{K}} = \left[ \mathcal{M}_\mathbf{K}; \mathbf{K}(x) \right], \quad \tilde{\mathbf{V}} = \left[ \mathcal{M}_\mathbf{V}; \mathbf{V}(x) \right] \\
        \text{Attention}(\mathbf{Q}, \tilde{\mathbf{K}}, \tilde{\mathbf{V}}) = \frac{\mathbf{Q}\tilde{\mathbf{K}}\tr}{\sqrt{d}} \tilde{\mathbf{V}}
    \end{gathered}
\end{equation}
where, for convenience, the exclusive dependency between regular keys and values and the current input sample $x$ has been made explicit, $\left[ \cdot; \cdot \right]$ indicates concatenation, $\mathcal{M}_\mathbf{K}$ the set of memory keys and $\mathcal{M}_\mathbf{V}$ the set of memory values.

Previous works which employed memory augmentation~\cite{sukhbaatar2019augmenting,cornia2020meshed,qi2021latent,chen2022reltransformer,xue2022mmt} have treated $\mathcal{M}_\mathbf{K}$ and $\mathcal{M}_\mathbf{V}$ as learnable parameters and, thus, optimized them directly through SGD during the learning process. Importantly, this imposes a constraint on what can be stored in memory vectors, as memories will be the result of accumulating gradient averaged over sequential mini-batches. This encourages the storage of information which is averagely beneficial to the entire training set and prevents focusing on the peculiarities of the single training items. As a consequence, learning a proper set of disentangled memory vectors turns out to be non-trivial and initialization-dependent~\cite{sukhbaatar2019augmenting}.

\subsection{Memories as banks of past activations}
We redefine memory keys and values as a means to let the network attend previous activations produced while processing other training samples. The network, at test time, will be able to attend to its own (past) activations produced while processing similar samples, thus aiding the generation process. Conceptually speaking, we might see this as a more principled design of a \textit{memory}, as in our case memory vectors will be actually storing past experiences of the network instead of being plain learnable parameters.

In our architecture, which we name \ours, we apply memories in a core position of the encoder-decoder structure, \ie~inside each self-attention layer of the captioner. This is different from what has been done in previous works (\eg~\cite{cornia2020meshed} considered only the Transformer encoder), and also represents a privileged placement for vision-and-language architectures, as the self-attention layer is in charge of modeling the temporal consistency of the generation and of integrating it with the result of the previous cross-attention layer, which connects with the visual modality. Figure~\ref{fig:model} (left) presents an overview of this design.

Considering a stream of mini-batches $\left[\mathbf{x}_0, \mathbf{x}_1, ..., \mathbf{x}_t, ...\right]$ containing randomly sampled training items, for each layer we define two memory banks $\mathcal{B}_\mathbf{K}$, $\mathcal{B}_\mathbf{V}$ which store all keys and values produced from past training samples, up to a maximum temporal distance of $\rm T$ iterations. Intuitively, the two memory banks model the manifold of keys and values seen over past training iterations. We then define the set of memory keys and values to be employed at the $t$-th training iteration as a function of the vectors contained in the respective memory banks:
\begin{equation}
    \label{eq:banks}
    \begin{gathered}
        \mathcal{B}_\mathbf{K} = \left[ \mathbf{K}(\mathbf{x}_{t-1}), \mathbf{K}(\mathbf{x}_{t-2}), ..., \mathbf{K}(\mathbf{x}_{t-\rm T}) \right], \\ 
        \mathcal{B}_\mathbf{V} = \left[ \mathbf{V}(\mathbf{x}_{t-1}), \mathbf{V}(\mathbf{x}_{t-2}), ..., \mathbf{V}(\mathbf{x}_{t-\rm T}) \right], \\ 
        \mathcal{M}_\mathbf{K} = f\left( \mathcal{B}_\mathbf{K} \right), \quad
        \mathcal{M}_\mathbf{V} = f\left( \mathcal{B}_\mathbf{V} \right).
    \end{gathered}
\end{equation}
In the equations above, for ease of notation, we denote with $\mathbf{K}(\mathbf{x})$ the set of keys produced by a layer while processing all items contained in a mini-batch $\mathbf{x}$. In practice, the temporal window $\rm T$ should be chosen to be sufficiently large to reasonably model the training set distribution (as shown in Sec.~\ref{sec:ablations}).
Also, memory banks need to be updated frequently and in a sufficiently smooth manner, so to follow the evaluation of the keys and values manifold and not to alter the training process, as will be discussed in the following.

\tit{Building memory prototypes}
Naively placing all keys and values produced during a given time window in the memory (\ie~setting $f(\cdot)$  to the identity in Eq.~\ref{eq:banks}) would require, approximately, $\rm T \cdot B \cdot h \cdot \tau$ memory slots per layer, where $\rm T$ represents the number of iterations executed inside the time window, $B$ the mini-batch size, $h$ the number of heads, and $\tau$ the average ground-truth sequence length. Under this setting, storing an entire COCO epoch would require storing around 96M memory vectors to both key and value sequences\footnote{Considering a network with 8 heads, and BPE tokenization.}, which would make the problem intractable in terms of memory occupation and computational complexity, because of the additional memory required to store vectors and because of the resulting growth of the attention matrix size. Further, as keys and values have been trained to summarize the information contained in a token with respect to other tokens of the same sequence, multiple elements in the memory bank could produce similar attention scores, thus increasing the entropy of the resulting attention distributions.

For this reason, we instead build synthetic key/value pairs as \textit{prototypical memory vectors} which are representative of the distribution of the entire memory bank. In doing this, we satisfy two design requirements:
(1) building prototypes should be fast, as we will be performing this on every layer of the architecture and several times during training; (2) memory prototypes should evolve during training to adapt to the changing distribution of keys and values.

In our method, prototype key memory vectors are obtained by clustering the manifold identified by the memory bank of keys and taking the resulting centroids. Value memory vectors are, instead, computed by interpolating between the values corresponding to the keys that lie in each cluster. Formally, being $m$ the target size of the memory, key memory vectors are obtained as follows:
\begin{equation}
    \label{eq:prototype_gen_k}
    \mathcal{M}_\mathbf{K} = \left[ \mathcal{M}_\mathbf{K}^1, \mathcal{M}_\mathbf{K}^2, ..., \mathcal{M}_\mathbf{K}^m \right] = \texttt{K-Means}_m(\mathcal{B}_\mathbf{K}),
\end{equation}
where function $\texttt{K-Means}_m(\cdot)$ returns the $m$ centroids obtained by performing a K-Means clustering over the key memory bank. Value memory vectors are computed by taking a linear combination of vectors in the value manifold that correspond to keys that lie close to key prototypes $\mathcal{M}_\mathbf{K}^i$, according to a distance function $d(\cdot)$ which compares items in the key manifold:
\begin{equation}
    \label{eq:prototype_gen_v}
    \begin{gathered}
        \mathcal{M}_\mathbf{V} = \left[ \mathcal{M}_\mathbf{V}^1, \mathcal{M}_\mathbf{V}^2, ..., \mathcal{M}_\mathbf{V}^m \right], \\
        \mathcal{M}_\mathbf{V}^i = \sum_{(\mathbf{K}^j, \mathbf{V}^j) \in \texttt{top-k}(\mathcal{M}_\mathbf{K}^i)} e^{-d\left(\mathcal{M}_\mathbf{K}^i, \mathbf{K}^j\right)} \mathbf{V}^j,
    \end{gathered}
\end{equation}
where, here, function $\texttt{top-k}(\mathcal{M}_\mathbf{K}^i)$ returns the closest (key, value) pairs in the memory bank with respect to $\mathcal{M}_\mathbf{K}^i$.

It shall be noted that the $\texttt{top-k}$ operation can be implemented by fitting a k-NN index on the keys memory bank ($\mathcal{B}_\mathbf{K}$) and using the centroid $\mathcal{M}_\mathbf{K}^i$ as query. The resulting list of keys close to $\mathcal{M}_\mathbf{K}^i$ can then be paired with their corresponding values to compute Eq.~\ref{eq:prototype_gen_v}. In practice, we use the $L_2$ distance as distance function inside both the key and value manifolds, as we found it to perform favorably compared to the inner product during preliminary experiments.

\tit{Discussion} With the strategy defined above, we obtain a set of $m$ memory keys and values, where $m$ can be controlled a-priori. Taking prototypes as centroids ensures, when $m$ is sufficiently high, that memory keys model the memory bank distribution properly. Further, the distance between centroids and cluster members in the key manifold is small, which has a positive effect on the resulting attention distribution, compared to the one obtainable by setting $f(\cdot)$ to the identity -- \ie, when storing all keys and values from the memory bank in the self-attention layer. Similar keys in an $L_2$ space, indeed, result in similar attention distributions, as we show in the following.

\noindent \textit{Proposition} -- Given a query $q$ and a set of keys $\mathbf{K}$, if a key $k \in \mathbf{K}$ is replaced with $\tilde{k}$ such that $\| k - \tilde{k} \|_2 \leq \varepsilon$ to form $\tilde{\mathbf{K}}$, then $\| \softmax(q\mathbf{K}\tr) - \softmax(q\tilde{\mathbf{K}}\tr) \|_2 \leq \varepsilon \| q \|_2$.

\noindent \textit{Proof.} As the $\softmax$ operator has Lipschitz constant less than 1~\cite{vyas2020fast,gao2017properties} and because the $L_2$ matrix norm is subordinate, $\| \softmax(q\mathbf{K}\tr) - \softmax(q\tilde{\mathbf{K}}\tr) \|_2 \leq \| q\mathbf{K}\tr - q\tilde{\mathbf{K}}\tr \|_2 \leq \| q \|_2 \| (\mathbf{K} - \tilde{\mathbf{K}})\tr \|_2 $. Recalling that, for any matrix $A$, $\| A \|_2 \leq \| A \|_F$, $\| (\mathbf{K} - \tilde{\mathbf{K}})\tr \|_2 \leq \| k - \tilde{k} \|_2$, from which the thesis follows. \qed

\begin{figure}[t]
\begin{center}
\includegraphics[width=\linewidth]{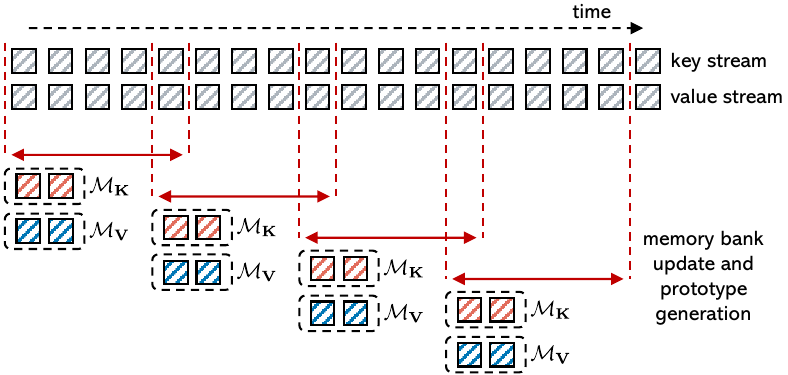}
\end{center}
\vspace{-.15cm}
\caption{Memory bank update approach.}
\label{fig:deque}
\vspace{-.4cm}
\end{figure}

\tit{Memory bank update} To ensure that memories are refreshed during training while lowering the total cost of prototypes generation, we adopt a strided sliding window approach to update the memory banks, which is visually depicted in Fig.~\ref{fig:deque}. Having defined a maximum length $\rm T$ for the banks (Eq.~\ref{eq:banks}), at regular intervals we take the last $\rm T$ batches from the key/value stream produced by a layer, create a memory bank with those and generate prototype vectors to be placed in $\mathcal{M}_\mathbf{K}$ and $\mathcal{M}_\mathbf{V}$ (Eq.~\ref{eq:prototype_gen_k},~\ref{eq:prototype_gen_v}). In practice, the process is repeated twice per epoch and memory banks store around two epochs of samples, so to have a significant overlap between the memory banks obtained at two consecutive update steps, which helps to stabilize the training. This is also illustrated in pseudo-code in Algorithm~\ref{algo:xe_pseudocode}.

\tit{Segment embeddings} As the final set of keys of the layer is a concatenation of memory-specific and input-specific keys (Eq.~\ref{eq:memory_def}), we add two different, learnable, segment embeddings to $\mathbf{K}$ and $\mathcal{M}_\mathbf{K}$, to help the network distinguish between the two key types.

\tit{Computational complexity}
Computing memory prototypes requires executing a K-Means clustering over the key memory bank ($\rm T \cdot B \cdot h \cdot \tau$ datapoints, $m$ clusters) and a kNN search over the key memory bank (which contains the same number of items) and is executed every $s$ training steps, being $s$ the stride employed over the key/value stream to update the memory banks (Fig.~\ref{fig:deque}). Further, the addition of the $m$ memory vectors to a mini-batch having a sequence length of $T$ implies growing the attention matrix from $T \times T$ to $(T + m) \times T$. As prototype generation is only required at training time, the latter is the only cost that is added at test time with respect to a standard attention layer.

In practice, adding prototypical memories does not increase inference times significantly with respect to a naive Transformer as the increase of the attention matrix is well amortized by the GPU parallelism. During training, computing the K-Means clustering and the k-NN index for solving Eq.~\ref{eq:prototype_gen_v} requires around 10s with a V100 GPU every time the memory needs to be refreshed. As the memory occupied for this can be de-allocated after prototypes computation, we did not need to decrease the batch size with respect to a baseline with learnable memory vectors.

\begin{figure}[t]
\vspace{-0.3cm}
\centering
\begin{minipage}{\linewidth}
\begin{algorithm}[H]
   \caption{\ours pseudocode}
    \definecolor{codeblue}{rgb}{0.25,0.5,0.5}
    \lstset{
      basicstyle=\fontsize{7.2pt}{7.9pt}\ttfamily\bfseries,
      commentstyle=\fontsize{7.2pt}{7.9pt}\color{codeblue},
      keywordstyle=\fontsize{7.2pt}{7.9pt},
    }
\label{algo:xe_pseudocode}
\vspace{-0.12cm}
\begin{lstlisting}[language=python]
# m: number of prototypes
# T: maximum length of the memory bank
# stride: memory bank update stride
# bank_k, bank_v: key/value memory banks
bank_k = [], bank_v = []
for img, caption in dataloader:
  output, act_k, act_v = net(img, caption)
  bank_k.append(act_k)
  bank_v.append(act_v)
  if len(bank_k) == T:
    compute_prototypes(m, bank_k, bank_v) # Eq. 3, 4
  bank_k = bank_k[stride:]
  bank_v = bank_v[stride:]
  loss = loss_fn(output, caption)
  loss.backward()
\end{lstlisting}
\vspace{-0.15cm}
\end{algorithm}
\end{minipage}
\vspace{-0.45cm}
\end{figure}

\section{Experimental Evaluation}
\label{sec:experiments}
\subsection{Datasets and evaluation protocol}
We analyze the effectiveness of our \ours on the widely used COCO benchmark~\cite{lin2014microsoft} employing the splits defined in~\cite{karpathy2015deep}. We also evaluate on the COCO online test server composed of more than 40k images for which ground-truth captions are not publicly available.

Additionally, we perform experiments on robust COCO, a different split of the COCO dataset introduced in~\cite{lu2018neural} to verify sensitivity to object hallucination and nocaps~\cite{agrawal2019nocaps} for novel object captioning. The former dataset guarantees that object pairs mentioned in captions of different sets do not overlap (with 110,234, 3,915, and 9,138 images for training, validation, and test), while the latter contains images annotated with 10 human-written captions, that can be further divided in in-domain, near-domain and out-of-domain pairs depending on their nearness to COCO.

To evaluate our results, we employ all standard captioning metrics, namely BLEU~\cite{papineni2002bleu}, METEOR~\cite{banerjee2005meteor}, ROUGE~\cite{lin2004rouge}, CIDEr~\cite{vedantam2015cider}, and SPICE~\cite{spice2016}, and some more recent evaluation scores like BERT-S~\cite{zhang2019bertscore}, CLIP-S~\cite{hessel2021clipscore}, and PAC-S~\cite{sarto2023positive} in both their reference-free and reference-based versions. When evaluating our results on robust COCO, we also employ the CHAIR metric~\cite{rohrbach2018object} that measures which fraction of objects mentioned in the generated sentences is hallucinated (CHi) and the portion of sentences that includes a hallucinated object (CHs).

\subsection{Implementation details}
Both the encoder and decoder are constructed with $L=6$ Transformer layers, with a hidden size of 512 and 8 attention heads. Unless otherwise specified, we employ 1,024 memory vectors and a size of the memory banks $T$ equal to 1,500.
We use a CLIP~\cite{radford2021learning} ViT-L/14 image encoder over the input image. The rationale behind this choice is that they have higher quality, adaptability to different tasks, and lower computational load compared to detection-based ones. To ensure fair comparison, we re-train recent and publicly-available models using the same features.

Our source code is based on Huggingface~\cite{wolf-etal-2020-transformers}, using the GPU-based implementations of K-Means and k-NN search from FAISS~\cite{johnson2019billion}. At training stage, the overall objective of \ours is the typical cross-entropy loss for sentence generation. Next, following~\cite{rennie2017self}, \ours can be further optimized with sentence-level reward, using the CIDEr score. Specifically, we first pre-train with the LAMB optimizer~\cite{you2019large}, a batch size of 1,024 and for 20,000 steps. We use the following learning rate schedule: we linearly warmup for 1,000 steps, then keep a constant learning rate of $2.5\cdot 10^{-4}$ until 10,000 steps, then sub-linearly decrease until 15,000 steps to $10^{-5}$ and keep the value constant for the rest of the training. For the second stage, we further optimize \ours using the Adam optimizer~\cite{kingma2015adam} and with $1\cdot 10^{-6}$ as learning rate, for 50,000 steps using a batch size of 64. We employ a beam size equal to 5.

\begin{table}[t]
\small
\centering
\setlength{\tabcolsep}{.3em}
\begin{center}
\resizebox{\linewidth}{!}{
\begin{tabular}{lccccccc}
\toprule
 & $m$ & $\rm T$ & B-4 & M & R & C & S \\
\midrule
Transformer~\cite{vaswani2017attention} & - & - & 37.4 & 30.3 & 58.9 & 127.8 & 23.3 \\
Transformer (w/ learnable mem.) & 64 & - & 37.7 & 30.2 & 58.1 & 127.9 & 23.4 \\
Transformer (w/ learnable mem.) & 1024 & - & 37.2 & 30.1 & 58.3 & 127.6 & 23.3 \\
\midrule
\ours & 256  & 1500 & 38.8 & 30.1 & 59.4 & 129.4 & 23.5 \\
\ours & 512  & 1500 & 39.0 & 30.1 & 59.5 & 130.0 & 23.5 \\
\ours & 1024 & 500  & 37.8 & 30.3 & 59.0 & 128.6 & 23.5 \\
\ours & 1024 & 1000 & 38.2 & \textbf{30.5} & 59.5 & 129.4 & 23.5 \\
\midrule
\ours (w/o mem. in 1st layer) & 1024 & 1500 & 38.3 & 30.2 & 59.0 & 129.2 & 23.3 \\
\ours (w/o segment emb.) & 1024 & 1500 & 38.6 & 30.4 & 59.4 & 130.1 & 23.4 \\
\rowcolor{blond}
\textbf{\ours} & 1024 & 1500 & \textbf{39.5} & 30.4 & \textbf{59.6} & \textbf{131.5} & \textbf{23.6} \\
\bottomrule
\end{tabular}
}
\end{center}
\vspace{-.15cm}
\caption{Ablation study ($m$ is the number of memory vectors and $\rm T$ is the size of the memory banks).}
\vspace{-.4cm}
\label{tab:ablation}
\end{table}

\begin{table*}[t]
\small
\centering
\setlength{\tabcolsep}{.5em}
\begin{center}
\resizebox{0.9\linewidth}{!}{
\begin{tabular}{lc ccc c cccccccc c cccccccc}
\toprule
& & \multicolumn{8}{c}{\textbf{Cross-Entropy Loss}} & & \multicolumn{8}{c}{\textbf{CIDEr Optimization}} \\
\cmidrule{3-10} \cmidrule{12-19}
& & B-1 & B-2 & B-3 & B-4 & M & R & C & S & & B-1 & B-2 & B-3 & B-4 & M & R & C & S \\
\midrule
Up-Down~\cite{anderson2018bottom} & & 77.2 & - & - & 36.2 & 27.0 & 56.4 & 113.5 & 20.3 & & 79.8 & - & - & 36.3 & 27.7 & 56.9 & 120.1 & 21.4 \\
GCN-LSTM~\cite{yao2018exploring} & & 77.3 & - & - & 36.8 & 27.9 & 57.0 & 116.3 & 20.9 & & 80.9 & - & - & 38.3 & 28.6 & 58.5 & 128.7 & 22.1 \\
SGAE~\cite{yang2019auto} & & 77.6 & - & - & 36.9 & 27.7 & 57.2 & 116.7 & 20.9 & & 81.0 & - & - & 39.0 & 28.4 & 58.9 & 129.1 & 22.2 \\ 
AoANet~\cite{huang2019attention} & & 77.4 & - & - & 37.2 & 28.4 & 57.5 & 119.8 & 21.3 & & 80.2 & - & - & 38.9 & 29.2 & 58.8 & 129.8 & 22.4 \\
$\mathcal{M}^2$ Transformer~\cite{cornia2020meshed} & & - & - & - & - & - & - & - & - & & 80.8 & - & - & 39.1 & 29.2 & 58.6 & 131.2 & 22.6 \\
X-Transformer~\cite{pan2020x} & & 77.3 & 61.5 & 47.8 & 37.0 & 28.7 & 57.5 & 120.0 & 21.8 & & 80.9 & 65.8 & 51.5 & 39.7 & 29.5 & 59.1 & 132.8 & 23.4 \\
DLCT~\cite{luo2021dual} & & - & - & - & - & - & - & - & - & & 81.4 & - & - & 39.8 & 29.5 & 59.1 & 133.8 & 23.0 \\
RSTNet~\cite{zhang2021rstnet} & & - & - & - & - & - & - & - & - & & 81.8 & - & - & 40.1 & 29.8 & 59.5 & 135.6 & 23.3 \\
DIFNet~\cite{wu2022difnet} & & - & - & - & - & - & - & - & - & & 81.7 & - & - & 40.0 & 29.7 & 59.4 & 136.2 & 23.2 \\
CaMEL~\cite{barraco2022camel} & & 78.3 & - & - & 39.1 & 29.4 & 58.5 & 125.7 & 22.2 & & 82.8 & - & - & 41.3 & 30.2 & 60.1 & 140.6 & 23.9 \\
COS-Net~\cite{li2022comprehending} & & \textbf{79.2} & 63.8 & 50.2 & 39.2 & 29.7 & 58.9 & 127.4 & 22.7 & & 82.7 & 68.2 & 54.0 & 42.0 & \textbf{30.6} & 60.6 & 141.1 & \textbf{24.6} \\
GRIT$^\ast$~\cite{nguyen2022grit} & & \textcolor{gray}{-} & \textcolor{gray}{-} & \textcolor{gray}{-} & \textcolor{gray}{-} & \textcolor{gray}{-} & \textcolor{gray}{-} & \textcolor{gray}{-} & \textcolor{gray}{-} & & \textcolor{gray}{84.2} & \textcolor{gray}{-} & \textcolor{gray}{-} & \textcolor{gray}{42.4} & \textcolor{gray}{30.6} & \textcolor{gray}{60.7} & \textcolor{gray}{144.2} & \textcolor{gray}{24.3} \\
\midrule
Transformer$^\dagger$ & & 76.4 & 61.0 & 47.9 & 37.4 & 30.3 & 58.9 & 127.8 & 23.3 & & 83.4 & 68.6 & 54.2 & 42.0 & 30.0 & 60.6 & 140.3 & 23.5 \\
$\mathcal{M}^2$ Transformer$^\dagger$~\cite{cornia2020meshed} & & 78.8 & 63.3 & 49.5 & 38.7 & 29.6 & 58.9 & 127.8 & 23.3 & & 83.7 & 69.2 & 54.8 & 42.3 & 30.5 & 61.0 & 141.2 & 23.6 \\
CaMEL$^\dagger$~\cite{barraco2022camel} & & 78.8 & 63.5 & 50.3 & 39.2 & 30.0 & 59.3 & 129.9 & 23.4 & & 83.6 & 69.0 & 54.7 & 42.4 & \textbf{30.6} & 60.9 & 142.4 & 23.6 \\
\rowcolor{blond}
\textbf{\ours} & & 79.0 & \textbf{64.2} & \textbf{50.7} & \textbf{39.5} & \textbf{30.4} & \textbf{59.6} & \textbf{131.5} & \textbf{23.6} & & \textbf{83.8} & \textbf{69.3} & \textbf{55.0} & \textbf{43.0} & \textbf{30.6} & \textbf{61.1} & \textbf{144.1} & 24.0 \\
\bottomrule
\end{tabular}
}
\end{center}
\vspace{-.15cm}
\caption{Comparison with the state of the art on the Karpathy test. The $\dagger$ marker indicates models re-trained with the same visual features used by our approach, while $\ast$ indicates finetuning of the visual backbone.}
\vspace{-.35cm}
\label{tab:sota_results}
\end{table*}

\subsection{Comparisons and ablation studies}
\label{sec:ablations}
We firstly conduct an ablation study to investigate how each design choice in our \ours influences the overall performances on COCO dataset. Table~\ref{tab:ablation} details the performance comparisons among different ablated runs. Note that all the results reported here are without self-critical training strategy.
We start from a base Transformer encoder-decoder architecture, which is also a degraded version of \ours without memory banks and prototype vectors.
Subsequently, we compare by adding learnable memory vectors as defined in~\cite{sukhbaatar2019augmenting,cornia2020meshed} but in the same position of \ours, \ie~in place of the self-attention layer in the sentence decoder instead of the visual encoder.
Then, we add memory banks and prototype vectors and vary the number of clusters and the size of the memory banks. 

\begin{table}[t]
\footnotesize
\begin{center}
\setlength{\tabcolsep}{.3em}
\resizebox{\linewidth}{!}{
\begin{tabular}{lc ccccc}
\toprule
 & Training & BERT-S & CLIP-S & RefCLIP-S & PAC-S & RefPAC-S \\
\midrule
Transformer$^\dagger$ & XE & 0.947 & 0.741 & 0.804 & 0.819 & 0.865 \\
$\mathcal{M}^2$ Transformer$^\dagger$~\cite{cornia2020meshed} & XE & 0.946 & 0.744 & 0.806 & 0.815 & 0.864 \\
CaMEL$^\dagger$~\cite{barraco2022camel} & XE & 0.947 & 0.745 & 0.807 & 0.818 & 0.865 \\
\rowcolor{blond}
\textbf{\ours} & XE & \textbf{0.948} & \textbf{0.754} & \textbf{0.812} & \textbf{0.821} & \textbf{0.868} \\
\midrule
Transformer$^\dagger$ & SCST & \textbf{0.947} & 0.749 & 0.807 & 0.818 & 0.864 \\
$\mathcal{M}^2$ Transformer$^\dagger$~\cite{cornia2020meshed} & SCST & 0.946 & 0.749 & 0.809 & 0.817 & 0.865 \\
CaMEL$^\dagger$~\cite{barraco2022camel} & SCST & 0.945 & 0.751 & 0.810 & 0.818 & 0.865 \\
\rowcolor{blond}
\textbf{\ours} & SCST & 0.946 & \textbf{0.755} & \textbf{0.814} & \textbf{0.821} & \textbf{0.869} \\
\bottomrule
\end{tabular}
}
\end{center}
\vspace{-.15cm}
\caption{Comparison with additional metrics. $\dagger$ indicates models re-trained with the same visual features used by our approach.}
\vspace{-.4cm}
\label{tab:otherscores}
\end{table}

Firstly, we notice that the basic learnable memory vectors do not give a significant contribution when placed in the sentence decoder, outlining that in this core position of the captioner, in which activations coming from both modalities are merged, learning appropriate memory vectors becomes more complex. Instead, the proposed prototype vectors increase caption quality significantly, up to 3.7 CIDEr points, highlighting the appropriateness of the proposed strategy. We notice that increasing the number of clusters and the size of the memory banks exhibits better performances, as we hypothesize that this provides a better estimation of the key and value manifold and more fine-grained prototypes. 

In our architecture, the sliding window contains the last $\rm T \cdot B$ captions seen, which in our best configuration amounts to 1.5M samples. Being COCO 0.6M image-text pairs, this models the training set distribution and its evolution across more than two epochs. We notice that increasing $\rm T$ further does not enhance performance; reducing it, especially to less than one epoch, is instead detrimental.

In the lower part of Table~\ref{tab:ablation}, we run additional ablations on two design choices: the use of segment embeddings to distinguish prototypes from input-dependent keys, and the incorporation of prototypical vectors in the first layer of the captioner. The second experiment arises from the fact that the first layer is not influenced by cross-attention results and, therefore, by multimodal connections with the input image. It can be observed that the segment embeddings provide a relevant contribution, and that memory prototypes have an impact both on the first layer and on the other layers, outlining that its advantage is inherently multimodal.

\begin{figure}[t]
\vspace{-.1cm}
\centering
\begin{center}
\includegraphics[width=.8\linewidth]{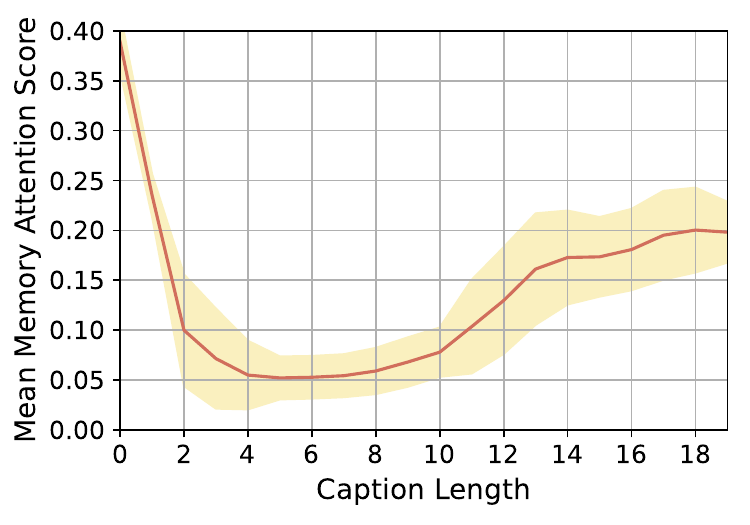}
\end{center}
\vspace{-.45cm}
\caption{Average magnitude of attention on prototype memories over time on the COCO Karpathy validation split.}
\label{fig:average_memory_attention_score}
\vspace{-.4cm}
\end{figure}

\tit{Memory attention visualization}
We show how the prototype memories are employed during the generation of the captions. To this aim, we compute a memory attention score that represents the percentage of the attention that is applied on the memories when the model has to decide which token to generate. For each generated token and a layer $l$, we look at the attention scores with respect to past keys ($\bm{a}^l_p$) and the memory ($\bm{a}^l_m$). We then compute the memory attention score for the layer as $\text{mean}(\bm{a}^l_m) / (\text{mean}(\bm{a}^l_m) + \text{mean}(\bm{a}^l_p))$, then average across layers. Figure~\ref{fig:average_memory_attention_score} shows the average attention scores over the COCO test set. As can be seen, the memory is employed during the generation of the entire sentence. Specifically, we observe a strong peak in the initial part of the generation, in which we speculate that the network retrieves a-priori information from the memory. Attention scores further increase in the final part, while the network describes details and less relevant objects which can benefit from the retrieval of additional knowledge.

\subsection{Comparison with the State-of-the-Art}
We then compare \ours with different state-of-the-art approaches. The models we compare to include the classic Up-Down~\cite{anderson2018bottom} approach, GCN-LSTM~\cite{yao2018exploring} that uses graph convolutional networks to encode visual relationships, SGAE~\cite{yang2019auto} which is based on scene graphs, AoANet~\cite{huang2019attention}, X-Transformer~\cite{pan2020x} and RSTNet~\cite{zhang2021rstnet} that propose attention variants, DLCT~\cite{luo2021dual}, DIFNet~\cite{wu2022difnet} that combine features extracted from multiple backbones, COS-Net~\cite{li2022comprehending} which retrieves knowledge from an external base, and GRIT~\cite{nguyen2022grit} that uses a DETR-based detector and a grid feature network for image encoding, which are both finetuned during training. For fairness, we also compare with methods re-trained using our visual features, namely CaMEL~\cite{barraco2022camel} which employs a mean teacher learning approach, and the $\mathcal{M}^2$ Transformer~\cite{cornia2020meshed}.

\begin{figure}[t]
\begin{center}
    \begin{minipage}{0.28\linewidth}
        \includegraphics[width=0.92\linewidth]{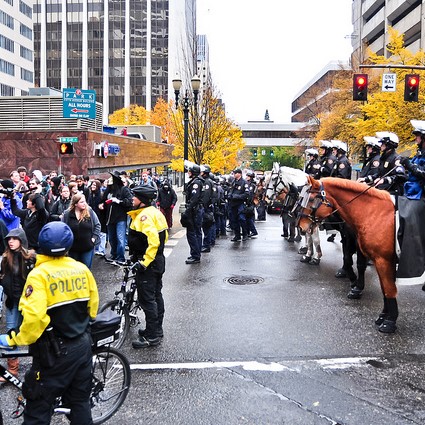}
        \end{minipage}
        \begin{minipage}{0.68\linewidth}
        \footnotesize{
        \textbf{GT:} A group of horse mounted police standing in front of a crowd.\\
        \textbf{Transformer:} A group of police officers standing in a street.\\
        \textbf{\ours:} A group of police officers on horses in a street.
        }
    \end{minipage}
    
    \vspace{0.07cm}
    
    \begin{minipage}{0.28\linewidth}
        \includegraphics[width=0.92\linewidth]{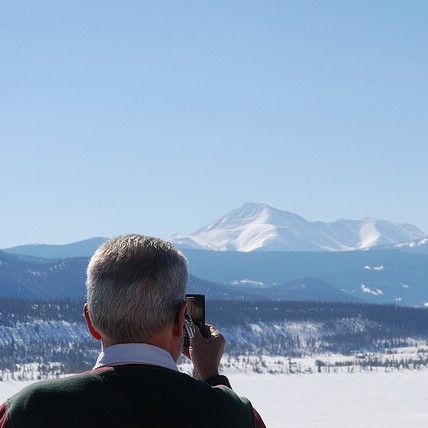}
        \end{minipage}
        \begin{minipage}{0.68\linewidth}
        \footnotesize{
        \textbf{GT:} A man takes a picture of snowy mountains with his cell phone.\\
        \textbf{Transformer:} A man taking a picture of the mountains.\\
        \textbf{\ours:} A man taking a picture of mountains with a cell phone.
        }
    \end{minipage}
    
    \vspace{0.07cm}
    
    \begin{minipage}{0.28\linewidth}
        \includegraphics[width=0.92\linewidth]{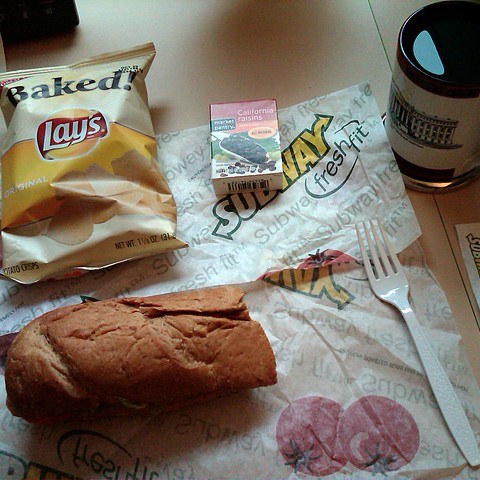}
        \end{minipage}
        \begin{minipage}{0.68\linewidth}
        \footnotesize{
        \textbf{GT:} A Subway sandwich with chips raisins and a coffee cup.\\
        \textbf{Transformer:} A sandwich and a bag of chips on a table.\\
        \textbf{\ours:} A table with a sandwich and chips and a cup of coffee.
        }
    \end{minipage}
\end{center}
\vspace{-.15cm}
\caption{Qualitative results on COCO sample images.}
\label{fig:qualitative_results}
\vspace{-.4cm}
\end{figure}

\begin{table*}[t]
\small
\centering
\setlength{\tabcolsep}{.53em}
\begin{center}
\resizebox{0.92\linewidth}{!}{
\begin{tabular}{lccccccccccccccccccccc}
\toprule
 & & \multicolumn{2}{c}{BLEU-1} & & \multicolumn{2}{c}{BLEU-2} & & \multicolumn{2}{c}{BLEU-3} & & \multicolumn{2}{c}{BLEU-4} & & \multicolumn{2}{c}{METEOR} & &  \multicolumn{2}{c}{ROUGE} & & \multicolumn{2}{c}{CIDEr} \\
\cmidrule{3-4} \cmidrule{6-7} \cmidrule{9-10} \cmidrule{12-13} \cmidrule{15-16} \cmidrule{18-19} \cmidrule{21-22} 
& & c5 & c40 & & c5 & c40 & & c5 & c40 & & c5 & c40 & & c5 & c40 & & c5 & c40 & & c5 & c40 \\
\midrule
Up-Down~\cite{anderson2018bottom} & & 80.2 & 95.2 & & 64.1 & 88.8 & & 49.1 & 79.4 & & 36.9 & 68.5 & & 27.6 & 36.7 & & 57.1 & 72.4 & & 117.9 & 120.5 \\
SGAE~\cite{yang2019auto} & & 81.0 & 95.3 & & 65.6 & 89.5 & & 50.7 & 80.4 & & 38.5 & 69.7 & & 28.2 & 37.2 & & 58.6 & 73.6 & & 123.8 & 126.5 \\
AoANet~\cite{huang2019attention} & & 81.0 & 95.0 & & 65.8 & 89.6 & & 51.4 & 81.3 & & 39.4 & 71.2 & & 29.1 & 38.5 & & 58.9 & 74.5 & & 126.9 & 129.6 \\
$\mathcal{M}^2$ Transformer~\cite{cornia2020meshed} & & 81.6 & 96.0 & & 66.4 & 90.8 & & 51.8 & 82.7 & & 39.7 & 72.8 & & 29.4 & 39.0 & & 59.2 & 74.8 & & 129.3 & 132.1 \\
X-Transformer~\cite{pan2020x} & & 81.9 & 95.7 & & 66.9 & 90.5 & & 52.4 & 82.5 & & 40.3 & 72.4 & & 29.6 & 39.2 & & 59.5 & 75.0 & & 131.1 & 133.5 \\
RSTNet~\cite{zhang2021rstnet} & & 82.1 & 96.4 & & 67.0 & 91.3 & & 52.2 & 83.0 & & 40.0 & 73.1 & & 29.6 & 39.1 & & 59.5 & 74.6 & & 131.9 & 134.0 \\
DLCT~\cite{luo2021dual} & & 82.4 & 96.6 & & 67.4 & 91.7 & & 52.8 & 83.8 & & 40.6 & 74.0 & & 29.8 & 39.6 & & 59.8 & 75.3 & & 133.3 & 135.4 \\
COS-Net~\cite{li2022comprehending} & & 83.3 & 96.8 & & 68.6 & 92.3 & & 54.2 & 84.5 & & 42.0 & 74.7 & & 30.4 & 40.1 & & 60.6 & 76.4 & & 136.7 & 138.3 \\
CaMEL~\cite{barraco2022camel} & & 83.2 & 97.3 & & 68.3 & 92.7 & & 53.6 & 84.8 & & 41.2 & 74.9 & & 30.2 & 39.7 & & 60.2 & 75.6 & & 137.5 & 140.0 \\
\midrule
\rowcolor{blond}
\textbf{\ours} & & \textbf{84.7} & \textbf{97.9} & & \textbf{70.2} & \textbf{93.8} & & \textbf{55.7} & \textbf{86.5} & & \textbf{43.4} & \textbf{77.1} & & \textbf{30.5} & \textbf{40.3} & & \textbf{61.3} & \textbf{76.8} & & \textbf{141.5} & \textbf{143.4} \\
\bottomrule
\end{tabular}
}
\end{center}
\vspace{-.15cm}
\caption{Leaderboard of various methods on the online COCO test server.
}
\vspace{-.35cm}
\label{tab:coco_test}
\end{table*}

\tit{Karpathy test split} In Table~\ref{tab:sota_results} we report results on the standard Karpathy test split, in a single-model setting. The upper part of the table shows the results reported by the compared approaches, using their original features. In the lower part, instead, we re-train different approaches on the same CLIP grid features we employ for training \ours. Specifically, in addition to a Transformer, we re-train the $\mathcal{M}^2$ Transformer~\cite{cornia2020meshed} and CaMEL~\cite{barraco2022camel}, which both represent recent and complementary approaches which could also be integrated with our proposal. With respect to a standard Transformer, \ours exhibits a margin of 3.8 CIDEr points also under CIDEr optimization, similarly to the XE training setting. Further, when compared with recent approaches using the same features, \ours also provides better performance. As shown in the table, with the exception of GRIT~\cite{nguyen2022grit} that differently from us finetunes the visual backbone, \ours consistently outperforms all the state-of-the-art approaches according to all metrics. Notably, our model is still competitive even compared to GRIT, despite the latter uses more powerful visual features. In Table~\ref{tab:otherscores}, we also compare against baselines trained on the same features using more recent learnable metrics, \ie~BERT-S, CLIP-S and, PAC-S. To qualitatively validate the effectiveness of our solution, we report sample images and corresponding predicted captions in Fig.~\ref{fig:qualitative_results}.

\tit{COCO Test Server} We also report the performances of our approach obtained on the official COCO test split, through the online test server\footnote{\scriptsize{\url{https://codalab.lisn.upsaclay.fr/competitions/7404}}}. Table~\ref{tab:coco_test} reports the performance with respect to 5 reference captions (c5) and 40 reference captions (c40). Following previous literature~\cite{cornia2020meshed,li2022comprehending}, we report the results using an ensemble of four models. As it can be seen, also in this setting \ours surpasses the compared approaches by a large margin, further demonstrating its effectiveness on the COCO dataset.

\begin{table}[t]
\small
\centering
\setlength{\tabcolsep}{.35em}
\begin{center}
\resizebox{0.96\linewidth}{!}{
\begin{tabular}{lc cccccccc}
\toprule
 & & B-1 & B-4 & M & R & C & S & CHs & CHi \\
\midrule
Att2In~\cite{rennie2017self} & & - & - & 24.0 & - & 85.8 & 16.9 & 14.1 & 10.1 \\
Up-Down~\cite{anderson2018bottom} & & - & - & 24.7 & - & 89.8 & 17.7 & 11.3 & 7.9 \\
Transformer\cite{li2022comprehending} & & 76.9 & 36.3 & 27.4 & 56.1 & 109.3 & 20.5 & 7.9 & 5.1 \\
COS-Net~\cite{li2022comprehending} & & 78.0 & 37.3 & 27.9 & 56.8 & 112.1 & 21.2 & 6.2 & 3.9 \\
\midrule 
Transformer$^\dagger$ & & 77.4 & 37.8 & \textbf{29.4} & 58.1 & 119.6 & 22.3 & 4.6 & 2.8 \\
\rowcolor{blond}
\textbf{\ours} & & \textbf{79.5} & \textbf{39.3} & \textbf{29.4} & \textbf{58.7} & \textbf{122.0} & \textbf{22.5} & \textbf{4.3} & \textbf{2.6} \\
\bottomrule
\end{tabular}
}
\end{center}
\vspace{-.15cm}
\caption{Results on robust COCO test set. The $\dagger$ marker indicates a model re-trained with the same visual features of our approach.}
\vspace{-.4cm}
\label{tab:robust}
\end{table}

\tit{Robust COCO split and sensitivity to hallucination}
As our approach relies on the memorization of other training samples, we verify whether the proposed strategy has an impact in terms of object hallucination. We perform this analysis by employing the robust COCO splits defined in~\cite{lu2018neural} and report the results in Table~\ref{tab:robust}, comparing with state-of-the-art approaches and with a Transformer trained with the same visual features. Both \ours and the Transformer baseline are re-trained from scratch on this dataset using cross-entropy loss only. In addition to standard evaluation metrics, we employ the CHAIR score, in its variants CHi and CHs, to measure object hallucination. From this analysis, it can be seen that the addition of prototypes memory vectors reduces the hallucination rate with respect to a Transformer, and that \ours performs favorably with respect to previous approaches also in this case. 

\begin{table}[t]
\small
\centering
\setlength{\tabcolsep}{.35em}
\begin{center}
\resizebox{0.88\linewidth}{!}{
\begin{tabular}{lc cc c cc c cc}
\toprule
 & & \multicolumn{2}{c}{In} & & \multicolumn{2}{c}{Out} & & \multicolumn{2}{c}{Overall} \\
\cmidrule{3-4} \cmidrule{6-7} \cmidrule{9-10}
& & C & S & & C & S & & C & S \\
\midrule
NBT~\cite{agrawal2019nocaps} & & 62.1 & 10.1 & & 62.4 & 8.9 & & 60.2 & 9.5 \\
Up-Down~\cite{agrawal2019nocaps} & & 80.0 & 12.0 & & 66.4 & 9.7 & & 73.1 & 11.1 \\
Transformer~\cite{cornia2020meshed} & & 78.0 & 11.0 & & 29.7 & 7.8 & & 54.7 & 9.8 \\
$\mathcal{M}^2$ Transformer~\cite{cornia2020meshed} & & 85.7 & 12.1 & & 38.9 & 8.9 & & 64.5 & 11.1 \\
GRIT$^\ast$~\cite{nguyen2022grit} & & \textcolor{gray}{105.9} & \textcolor{gray}{13.6} & & \textcolor{gray}{72.6} & \textcolor{gray}{11.1} & & \textcolor{gray}{90.2} & \textcolor{gray}{12.8} \\
\midrule
Transformer$^\dagger$ & & 105.9 & 13.3 & & 73.9 & 11.3 & & 90.9 & 12.6 \\
\rowcolor{blond}
\textbf{\ours} & & \textbf{107.5} & \textbf{13.7} & & \textbf{75.9} & \textbf{11.4} & & \textbf{92.6} & \textbf{12.8} \\
\bottomrule
\end{tabular}
}
\end{center}
\vspace{-.15cm}
\caption{Results on nocaps validation set. The $\dagger$ marker indicates a model re-trained with the same visual features of our approach, while $\ast$ indicates finetuning of the visual backbone.}
\label{tab:nocaps}
\vspace{-.4cm}
\end{table}

\tit{Novel object captioning}
We also evaluate \ours on the nocaps dataset~\cite{agrawal2019nocaps} for novel object captioning. It shall be noted that our approach does not leverage components which are explicitly designed to deal with the naming of novel objects, still the nocaps dataset provides a relevant test bed to compare \ours with other approaches from the literature. To conduct this analysis, we employ our model and the Transformer-based baseline trained on the standard COCO dataset. Results are reported in Table~\ref{tab:nocaps}, both in the in-domain and out-of-domain splits of the datasets and without employing constrained beam search~\cite{anderson2017guided}. We observe that \ours achieves the best performance among all the compared approaches and with respect to the base Transformer which does not employ memory prototypes. In this setting, \ours also outperforms the results of GRIT, which employs a finetuned visual backbone. This highlights that the addition of prototypes memory vectors improves the description of novel objects.

\section{Conclusion}
\label{sec:conclusion}
We presented \ours, a novel architecture for image captioning that is based on novel prototypical memory vectors which are integrated into standard attention layers to summarize activations produced during recent training iterations. Noticeably, the exploitation of past activations and the construction of memory prototypes is unprecedented in vision-and-language architectures. Experimental results demonstrate the effectiveness of \ours on the COCO dataset and show that it can alleviate hallucination effects and describe novel objects better than competitors. 

\section*{Acknowledgments}
\small{We thank CINECA for providing computational resources. This work has been supported by the PNRR-M4C2 project (PE00000013) ``FAIR - Future Artificial Intelligence Research'' funded by the European Commission and the PRIN ``CREATIVE: CRoss-modal understanding and gEnerATIon of Visual and tExtual content'' co-funded by the Italian Ministry of University and Research (CUP B87G22000460001).}

{\small
\bibliographystyle{ieee_fullname}
\bibliography{bibliography}
}

\newpage
\appendix
In the following, we present additional materials about \ours. In particular, we provide additional experimental results using model ensembling, visualizations of memory attention scores on sample images from the COCO dataset, and qualitative results on all considered datasets.

\section{Additional Experimental Results}
As a complement of the experiments reported in the main paper, in Table~\ref{tab:ensemble} we report the performance of \ours using an ensemble of four models on the Karpathy test split, in comparison with ensembles built with a base Transformer using the same visual features and with related methods from the literature. \ours confirms its effectiveness showing an improvement over the other competitors and the baseline version.

\section{Visualizations}
\begin{figure*}
\centering
\resizebox{\linewidth}{!}{
\includegraphics[width=\linewidth]{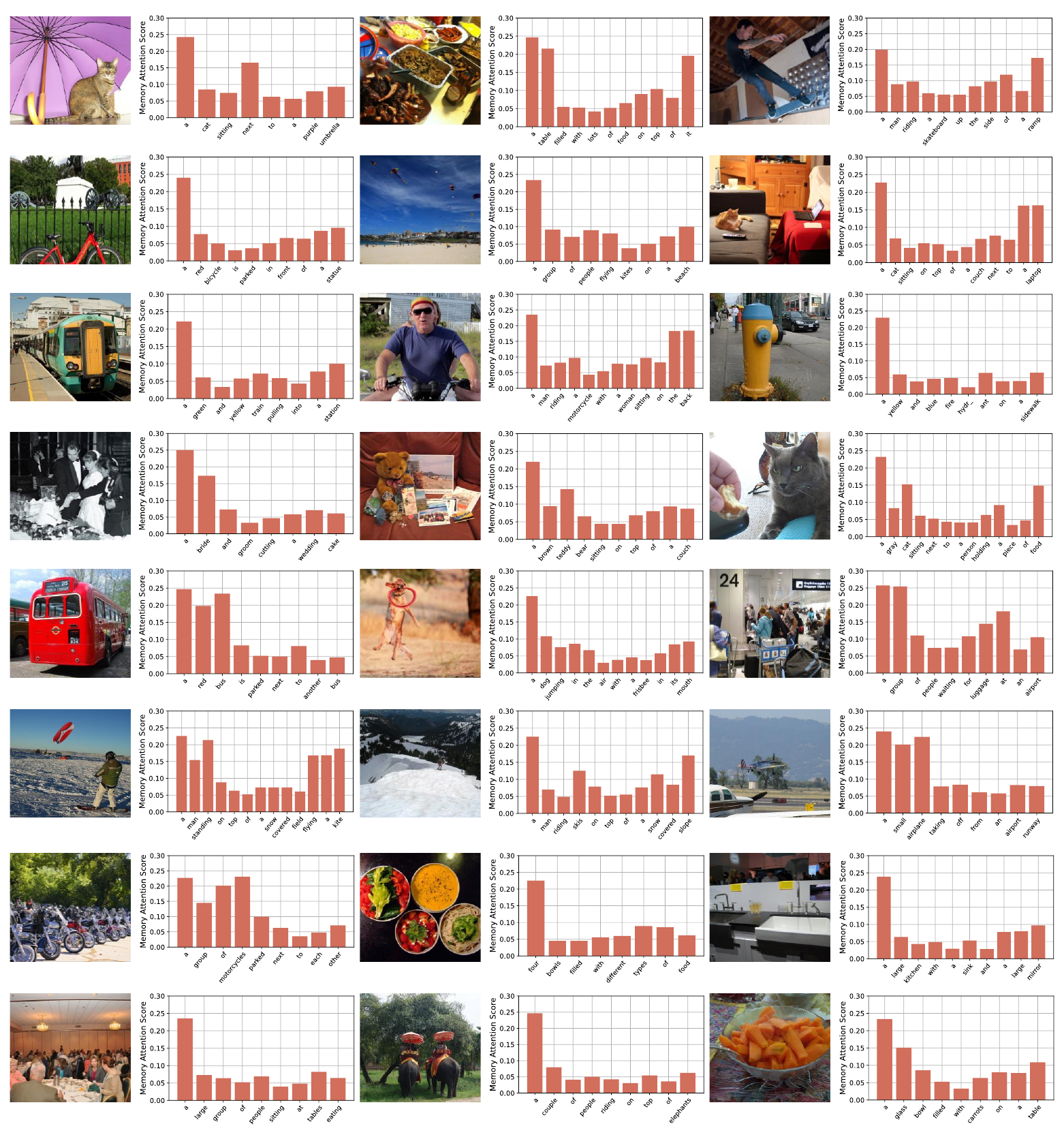}
}
\vspace{-.5cm}
\caption{Sample captions generated on the COCO dataset, together with the magnitude of attention on prototype memories over time.}
\label{fig:histograms}
\end{figure*}

In addition to the aggregate visualization of the attentive distributions shown in the main paper on the entire COCO test set, in Figure~\ref{fig:histograms} we show some examples of the usage of the prototypical memories on sample captions in a non-aggregated way. The attention score for each word is computed in the same manner: we calculate the relative average of the attention scores related to the memory slots over the totality of the attention scores for each layer, then we average over all the layers.
These visualizations show how memories are employed during caption generation in different scenarios, \ie~for retrieving further details on specific concepts, describing actions, and using more appropriate nouns and verbs with respect to other approaches. As also shown in the main paper, the prototype memories are in general employed during the generation of the entire sentence.

\begin{table}[b]
\small
\centering
\setlength{\tabcolsep}{.35em}
\begin{center}
\resizebox{\linewidth}{!}{
\begin{tabular}{lc cccccccc}
\toprule
& & B-1 & B-2 & B-3 & B-4 & M & R & C & S \\
\midrule
GCN-LSTM~\cite{yao2018exploring} & & 80.9 & - & - & 38.3 & 28.6 & 58.5 & 128.7 & 22.1 \\
SGAE~\cite{yang2019auto} & & 81.0 & - & - & 39.0 & 28.4 & 58.9 & 129.1 & 22.2 \\
AoANet~\cite{huang2019attention}  & & 81.6 & - & - & 40.2 & 29.3 & 59.4 & 132.0 & 22.8 \\
$\mathcal{M}^2$ Transformer~\cite{cornia2020meshed} & & 82.0 & - & - & 40.5 & 29.7 & 59.5 & 134.5 & 23.5 \\
X-Transformer~\cite{pan2020x} & & 81.7 & 66.8 & 52.6 & 40.7 & 29.9 & 59.7 & 135.3 & 23.8 \\
DLCT~\cite{luo2021dual} & & 82.2 & - & - & 40.8 & 29.9 & 59.8 & 137.5 & 23.3 \\
COS-Net~\cite{li2022comprehending} & & 83.5 & 69.1 & 54.9 & 42.9 & \textbf{30.8} & 61.0 & 143.0 & \textbf{24.7} \\
\midrule 
Transformer & & 83.9 & 69.1 & 54.6 & 42.3 & 30.2 & 60.7 & 141.3 & 23.6 \\
\rowcolor{blond}
\textbf{\ours} & & \textbf{84.9} & \textbf{70.4} & \textbf{56.1} & \textbf{43.8} & 30.7 & \textbf{61.6} & \textbf{145.9} & 24.1 \\
\bottomrule
\end{tabular}
}
\end{center}
\vspace{-.15cm}
\caption{Comparison with the state of the art on the COCO Karpathy test split using an ensemble of models.}
\vspace{-.3cm}
\label{tab:ensemble}
\end{table}

\section{Qualitative Results}
\begin{figure*}
\begin{center}
    \centering
    \begin{minipage}{0.152\linewidth}
        \includegraphics[width=0.95\linewidth]{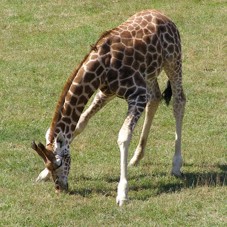}
        \end{minipage}
    \begin{minipage}{0.315\linewidth}
        \footnotesize{
        \textbf{Ground-truth:} A giraffe bending over while standing on green grass.\\
        \textbf{Transformer:} A giraffe bending down to eat the grass.\\
        \textbf{\ours:} A giraffe bending down to eat grass in a field.
        }
    \end{minipage}
    \hspace{0.3cm}
    \begin{minipage}{0.152\linewidth}
        \includegraphics[width=0.95\linewidth]{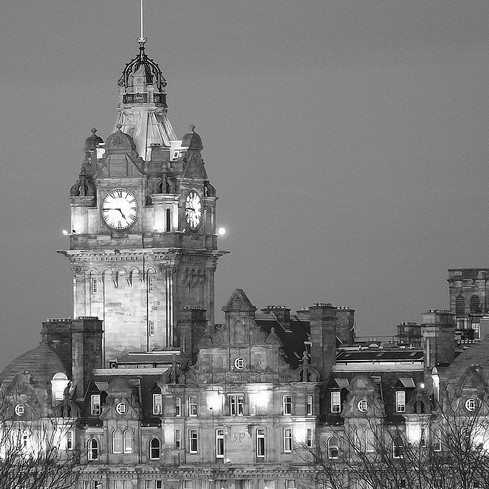}
        \end{minipage}
        \begin{minipage}{0.315\linewidth}
        \footnotesize{
        \textbf{Ground-truth:} A black and white photo of a castle at night.\\
        \textbf{Transformer:} A large building with a clock tower on top of.\\
        \textbf{\ours:} A black and white photo of a building with a clock.
        }
    \end{minipage}
                
    \vspace{0.20cm}
    
    \begin{minipage}{0.152\linewidth}
        \includegraphics[width=0.95\linewidth]{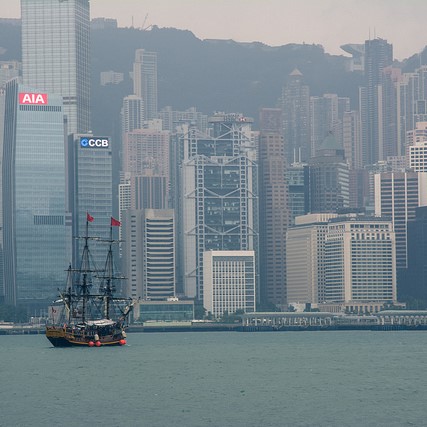}
        \end{minipage}
    \begin{minipage}{0.315\linewidth}
        \footnotesize{
        \textbf{Ground-truth:} A ship in the water sailing past the city in the background.\\
        \textbf{Transformer:} A boat in a large body of water.\\
        \textbf{\ours:} A boat in the water near a city.
        }
    \end{minipage}
    \hspace{0.3cm}
    \begin{minipage}{0.152\linewidth}
        \includegraphics[width=0.95\linewidth]{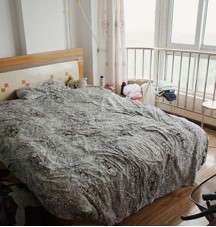}
        \end{minipage}
    \begin{minipage}{0.315\linewidth}
        \footnotesize{
        \textbf{Ground-truth:} A bed sitting on a hard wood floor.\\
        \textbf{Transformer:} A bedroom with a bed and a desk.\\
        \textbf{\ours:} A bedroom with a large bed and a window.
        }
    \end{minipage}
                    
    \vspace{0.20cm}
    
    \begin{minipage}{0.152\linewidth}
        \includegraphics[width=0.95\linewidth]{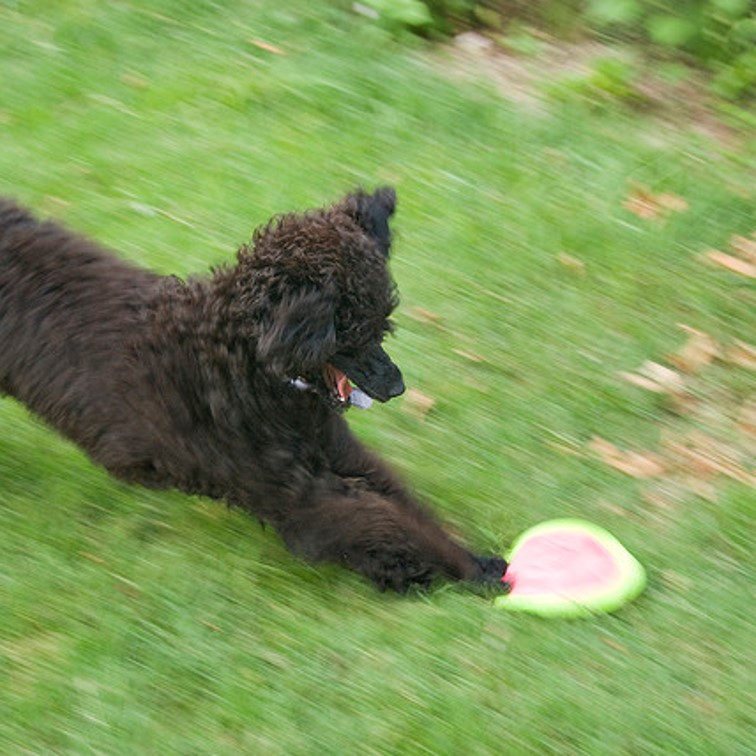}
        \end{minipage}
    \begin{minipage}{0.315\linewidth}
        \footnotesize{
        \textbf{Ground-truth:} A small black dog playing with a frisbee.\\
        \textbf{Transformer:} A black dog running with a frisbee in its mouth.\\
        \textbf{\ours:} A dog playing with a frisbee in the grass.
        }
    \end{minipage}
    \hspace{0.3cm}
    \begin{minipage}{0.152\linewidth}
        \includegraphics[width=0.95\linewidth]{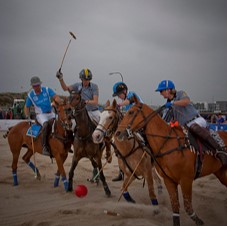}
        \end{minipage}
    \begin{minipage}{0.315\linewidth}
        \footnotesize{
        \textbf{Ground-truth:} Four people compete on horseback playing playing polo.\\
        \textbf{Transformer:} A group of men on horses playing soccer on a beach.\\
        \textbf{\ours:} A group of men playing polo on the beach.
        }
    \end{minipage}
                    
    \vspace{0.20cm}
    
    \begin{minipage}{0.152\linewidth}
        \includegraphics[width=0.95\linewidth]{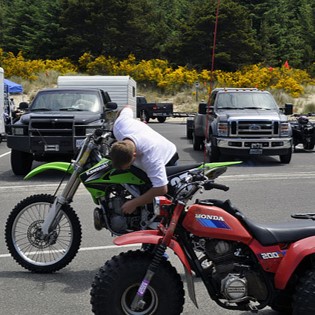}
        \end{minipage}
    \begin{minipage}{0.315\linewidth}
        \footnotesize{
        \textbf{Ground-truth:} A man bending down to fix his motorcycle in a parking lot.\\
        \textbf{Transformer:} A man on a dirt motorcycle on a road. \\
        \textbf{\ours:} A man leaning on a motorcycle in a parking lot.
        }
    \end{minipage}
    \hspace{0.3cm}
    \begin{minipage}{0.152\linewidth}
        \includegraphics[width=0.95\linewidth]{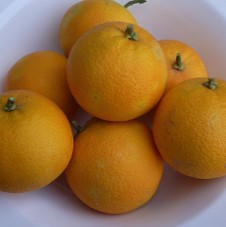}
        \end{minipage}
    \begin{minipage}{0.315\linewidth}
        \footnotesize{
        \textbf{Ground-truth:} Many oranges have been placed inside a bowl.\\
        \textbf{Transformer:} A white plate of four oranges on a table.\\
        \textbf{\ours:} A white bowl of oranges on a table.
        }
    \end{minipage}
                    
    \vspace{0.20cm}
    
    \begin{minipage}{0.152\linewidth}
        \includegraphics[width=0.95\linewidth]{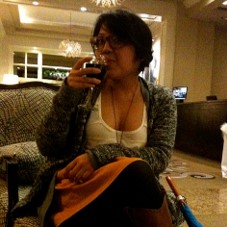}
        \end{minipage}
    \begin{minipage}{0.315\linewidth}
        \footnotesize{
        \textbf{Ground-truth:} A woman sitting in a chair and drinking from a wine glass.\\
        \textbf{Transformer:} A woman sitting on a couch drinking a glass.\\
        \textbf{\ours:} A woman sitting on a couch holding a glass of wine.
        }
    \end{minipage}
    \hspace{0.3cm}
    \begin{minipage}{0.152\linewidth}
        \includegraphics[width=0.95\linewidth]{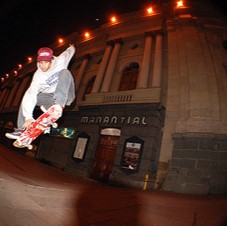}
        \end{minipage}
    \begin{minipage}{0.315\linewidth}
        \footnotesize{
        \textbf{Ground-truth:} A man with a skateboard that is jumping in the air.\\
        \textbf{Transformer:} A man is doing a trick on a skateboard.\\
        \textbf{\ours:} A man flying through the air while riding a skateboard.
        }
    \end{minipage}
                    
    \vspace{0.20cm}
    
    \begin{minipage}{0.152\linewidth}
        \includegraphics[width=0.95\linewidth]{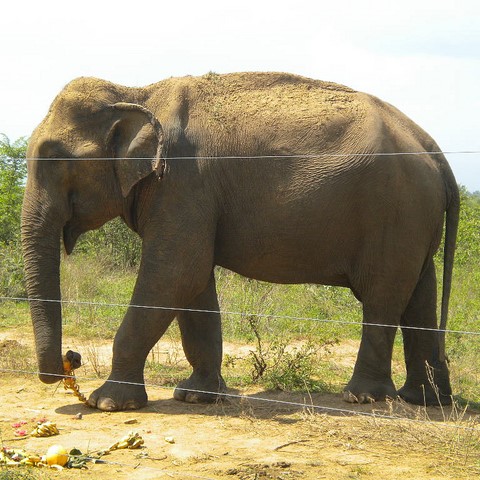}
        \end{minipage}
    \begin{minipage}{0.315\linewidth}
        \footnotesize{
        \textbf{Ground-truth:} An elephant can be seen through a barbed wire fence.\\
        \textbf{Transformer:} An elephant standing next to a wire fence.\\
        \textbf{\ours:} An elephant standing behind a barbed wire fence.
        }
    \end{minipage}
    \hspace{0.3cm}
    \begin{minipage}{0.152\linewidth}
        \includegraphics[width=0.95\linewidth]{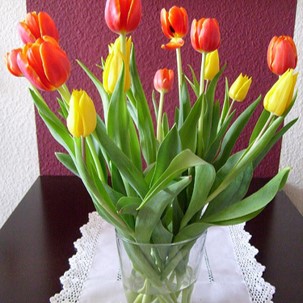}
        \end{minipage}
    \begin{minipage}{0.315\linewidth}
        \footnotesize{
        \textbf{Ground-truth:} A vase filled with yellow and red flowers.\\
        \textbf{Transformer:} A vase of flowers sitting on a table.\\
        \textbf{\ours:} A vase filled with red and yellow flowers on a table.
        }
    \end{minipage}
                        
    \vspace{0.20cm}
    
    \begin{minipage}{0.152\linewidth}
        \includegraphics[width=0.95\linewidth]{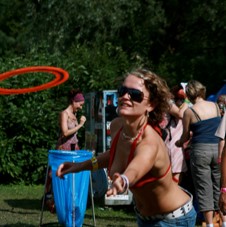}
        \end{minipage}
    \begin{minipage}{0.315\linewidth}
        \footnotesize{
        \textbf{Ground-truth:} A girl playing with a red frisbee outside at the park.\\
        \textbf{Transformer:} A woman throwing a frisbee in a park.\\
        \textbf{\ours:} A woman throwing a red frisbee in a park.
        }
    \end{minipage}
    \hspace{0.3cm}
    \begin{minipage}{0.152\linewidth}
        \includegraphics[width=0.95\linewidth]{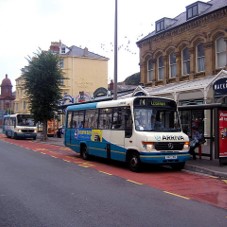}
        \end{minipage}
    \begin{minipage}{0.315\linewidth}
        \footnotesize{
        \textbf{Ground-truth:} A bus is parked near a bus stop on a street.\\
        \textbf{Transformer:} Two buses parked on the side of a street.\\
        \textbf{\ours:} A blue and white bus parked at a bus stop.
        }
    \end{minipage}
                        
    \vspace{0.20cm}
    
    \begin{minipage}{0.152\linewidth}
        \includegraphics[width=0.95\linewidth]{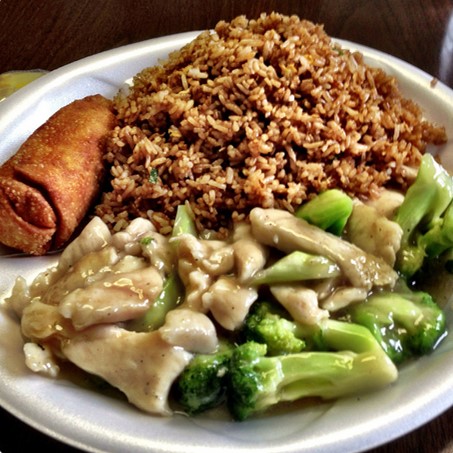}
        \end{minipage}
    \begin{minipage}{0.315\linewidth}
        \footnotesize{
        \textbf{Ground-truth:} Well cooked rice and vegetables on a white plate.\\
        \textbf{Transformer:} A plate of food with rice and rice on a.\\
        \textbf{\ours:} A plate of food with rice and broccoli on a table.
        }
    \end{minipage}
    \hspace{0.3cm}
    \begin{minipage}{0.152\linewidth}
        \includegraphics[width=0.95\linewidth]{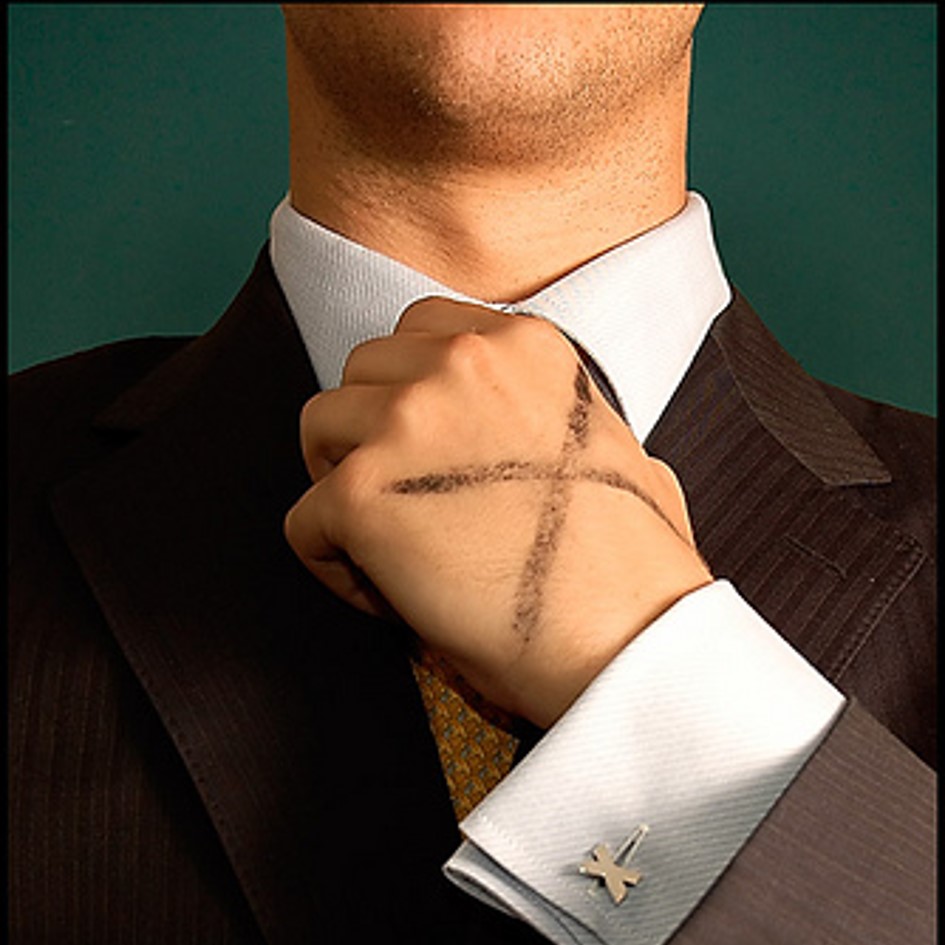}
        \end{minipage}
    \begin{minipage}{0.315\linewidth}
        \footnotesize{
        \textbf{Ground-truth:} A man in a suit carefully adjusts his tie.\\
        \textbf{Transformer:} A man in a suit tying his tie.\\
        \textbf{\ours:} A man adjusting his tie in a suit.
        }
    \end{minipage}
\end{center}
\caption{Qualitative results on sample images from the COCO Karpathy test split.}
\label{fig:coco_qualitatives}
\vspace{-.3cm}
\end{figure*}

\begin{figure*}[t]
\begin{center}
    \begin{minipage}{0.152\linewidth}
        \includegraphics[width=0.95\linewidth]{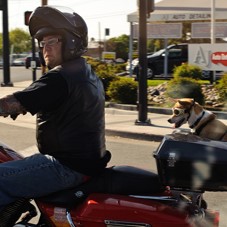}
        \end{minipage}
    \begin{minipage}{0.315\linewidth}
        \footnotesize{
        \textbf{Ground-truth:} A man riding a red motorcycle down a street with a dog on back of it.\\
        \textbf{Transformer:} A man riding on the back of a red motorcycle.\\
        \textbf{\ours:} A man riding a motorcycle with a dog on the back.
        }
    \end{minipage}
    \hspace{0.3cm}
    \begin{minipage}{0.152\linewidth}
        \includegraphics[width=0.95\linewidth]{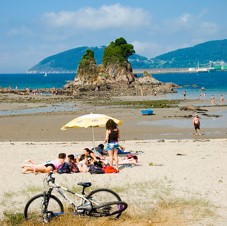}
        \end{minipage}
        \begin{minipage}{0.315\linewidth}
        \footnotesize{
        \textbf{Ground-truth:} The group of people lay on the beach near a parked bicycle.\\
        \textbf{Transformer:} A group of people sitting on top of a sandy beach.\\
        \textbf{\ours:} A group of people sitting on a beach next to a bike.
        }
    \end{minipage}
                
    \vspace{0.20cm}
    
    \begin{minipage}{0.152\linewidth}
        \includegraphics[width=0.95\linewidth]{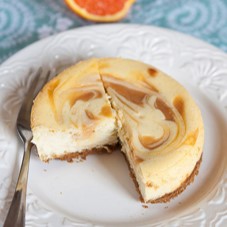}
        \end{minipage}
    \begin{minipage}{0.315\linewidth}
        \footnotesize{
        \textbf{Ground-truth:} A piece of dessert and fork are on the plate.\\
        \textbf{Transformer:} A slice of pie sitting on top of a white plate.\\
        \textbf{\ours:} A slice of cheesecake on a plate with a fork.
        }
    \end{minipage}
    \hspace{0.3cm}
    \begin{minipage}{0.152\linewidth}
        \includegraphics[width=0.95\linewidth]{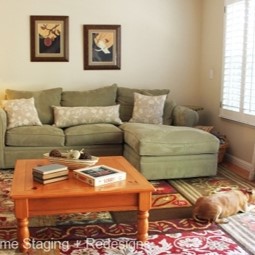}
        \end{minipage}
    \begin{minipage}{0.315\linewidth}
        \footnotesize{
        \textbf{Ground-truth:} A dog is lying on the carpet of the living room.\\
        \textbf{Transformer:} A living room with couches and a coffee table.\\
        \textbf{\ours:} A living room filled with furniture and a dog laying on a rug.
        }
    \end{minipage}
                    
    \vspace{0.20cm}
    
    \begin{minipage}{0.152\linewidth}
        \includegraphics[width=0.95\linewidth]{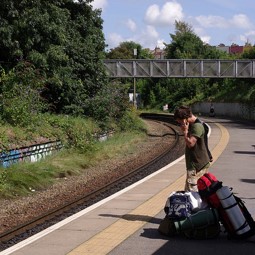}
        \end{minipage}
    \begin{minipage}{0.315\linewidth}
        \footnotesize{
        \textbf{Ground-truth:} A man on a cell phone waits with luggage by a train track.\\
        \textbf{Transformer:} A man standing on a train platform next to luggage.\\
        \textbf{\ours:} A man talking on a cell phone while standing next to a train track.
        }
    \end{minipage}
    \hspace{0.3cm}
    \begin{minipage}{0.152\linewidth}
        \includegraphics[width=0.95\linewidth]{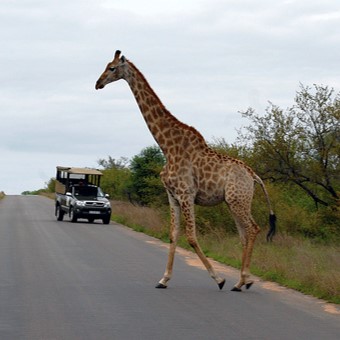}
        \end{minipage}
    \begin{minipage}{0.315\linewidth}
        \footnotesize{
        \textbf{Ground-truth:} A giraffe is crossing the street in front of a car.\\
        \textbf{Transformer:} A giraffe walking down a road next to a truck.\\
        \textbf{\ours:} A giraffe crossing a road next to a vehicle.
        }
    \end{minipage}
\end{center}
\caption{Qualitative results on sample images from the robust COCO test set.}
\label{fig:robust_coco_qualitatives}
\vspace{-.3cm}
\end{figure*}

\begin{figure*}[t]
\begin{center}
    \centering
    \begin{minipage}{0.152\linewidth}
        \includegraphics[width=0.95\linewidth]{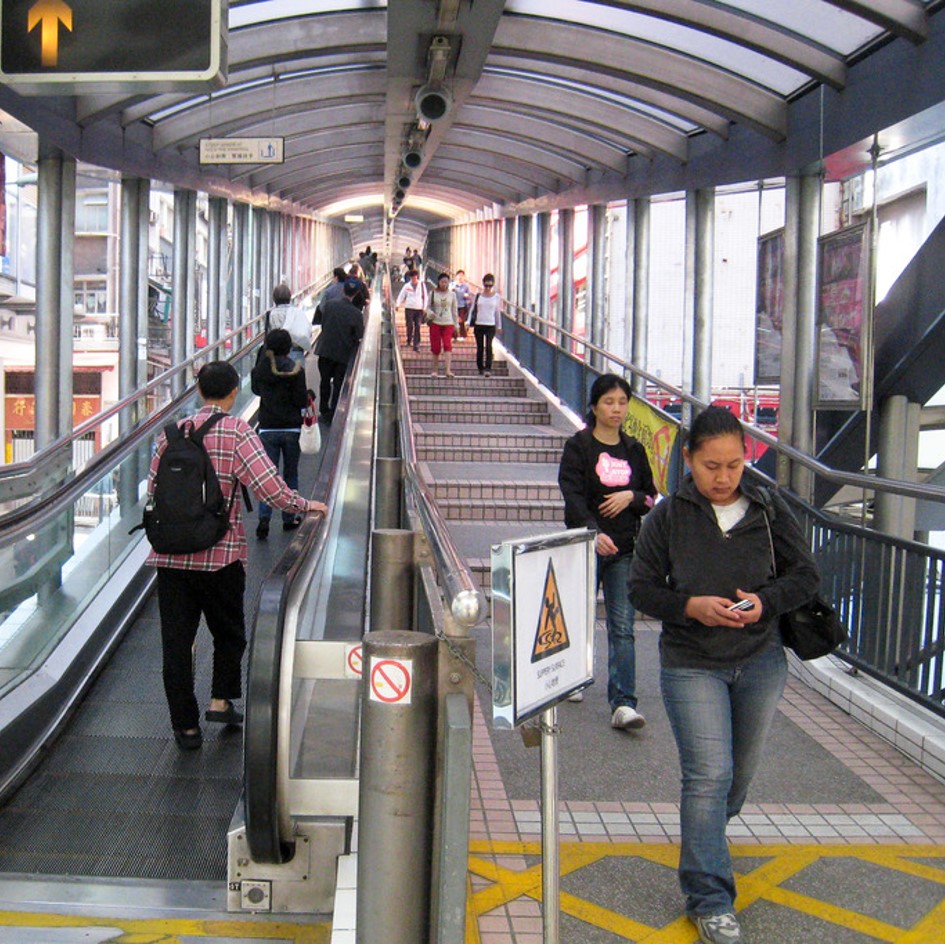}
        \end{minipage}
    \begin{minipage}{0.315\linewidth}
        \footnotesize{
        \textbf{Transformer:} A group of people walking down a subway train.\\
        \textbf{\ours:} A group of people walking up an escalator.
        }
    \end{minipage}
    \hspace{0.3cm}
    \begin{minipage}{0.152\linewidth}
        \includegraphics[width=0.95\linewidth]{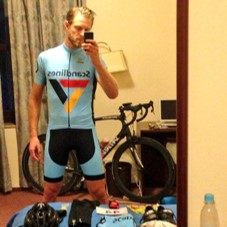}
        \end{minipage}
        \begin{minipage}{0.315\linewidth}
        \footnotesize{
        \textbf{Transformer:} A man is standing in front of a bike.\\
        \textbf{\ours:} A man taking a picture of himself in front of a mirror.
        }
    \end{minipage}
                
    \vspace{0.20cm}
    
    \begin{minipage}{0.152\linewidth}
        \includegraphics[width=0.95\linewidth]{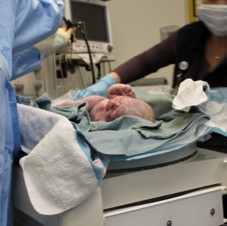}
        \end{minipage}
    \begin{minipage}{0.315\linewidth}
        \footnotesize{
        \textbf{Transformer:} A woman is washing a baby on a machine.\\
        \textbf{\ours:} A woman holding a baby in a hospital room.
        }
    \end{minipage}
    \hspace{0.3cm}
    \begin{minipage}{0.152\linewidth}
        \includegraphics[width=0.95\linewidth]{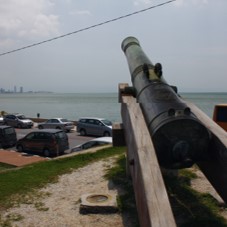}
        \end{minipage}
    \begin{minipage}{0.315\linewidth}
        \footnotesize{
        \textbf{Transformer:} A pirate ship on the water near a.\\
        \textbf{\ours:} A metal barrel sitting next to a body of water.
        }
    \end{minipage}
                    
    \vspace{0.20cm}
    
    \begin{minipage}{0.152\linewidth}
        \includegraphics[width=0.95\linewidth]{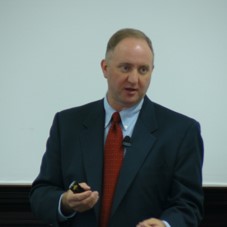}
        \end{minipage}
    \begin{minipage}{0.315\linewidth}
        \footnotesize{
        \textbf{Transformer:} A man in a suit and tie holding a microphone.\\
        \textbf{\ours:} A man in a suit and tie holding a remote.
        }
    \end{minipage}
    \hspace{0.3cm}
    \begin{minipage}{0.152\linewidth}
        \includegraphics[width=0.95\linewidth]{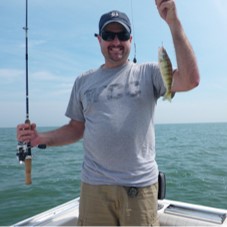}
        \end{minipage}
    \begin{minipage}{0.315\linewidth}
        \footnotesize{
        \textbf{Transformer:} A man is on a boat in the water.\\
        \textbf{\ours:} A man standing on a boat holding a fish.
        }
    \end{minipage}
\end{center}
\caption{Qualitative results on sample images from the nocaps validation set.}
\label{fig:nocaps_qualitatives}
\vspace{-.3cm}
\end{figure*}

Finally, we report qualitative results on different datasets. In particular, Figure~\ref{fig:coco_qualitatives} shows sample results from the COCO Karpathy test split, obtained from \ours and the baseline version, while additional samples are reported on the robust test split of COCO~\cite{lu2018neural} in Figure~\ref{fig:robust_coco_qualitatives} and on the validation split of the nocaps~\cite{agrawal2019nocaps} dataset in Figure~\ref{fig:nocaps_qualitatives}. 

We observe that, on average, \ours generates more detailed and correct descriptions with respect to a vanilla Transformer architecture. Further, the proposed approach showcases better performances when describing pairs of objects which do not appear in the training set and when describing out-of-domain objects.

\end{document}